\documentclass{article}
\usepackage{spconf,amsmath,amssymb,graphicx,hyperref}
\usepackage{booktabs}
\usepackage{booktabs}
\usepackage{multirow}
\usepackage{array}
\usepackage{graphicx}
\usepackage{capt-of}
\usepackage{xcolor}
\usepackage{caption}
\usepackage{amssymb}

\title{E2I-Flow: EEG-to-Image Reconstruction via Wavelet-Temporal Cross-Attention and Conditional Flow Matching}

\name{Haowei Pan$^{1}$, Hwa Hui Tew$^{1}$, Junn Yong Loo$^{2}$, Fuad Noman$^{1}$, Hernando Ombao$^{3}$, Chee-Ming Ting$^{1,*}$
\thanks{\mbox{\textsuperscript{*}Corresponding author: Chee-Ming Ting, Ting.CheeMing@monash.edu}}
}
\address{
$^{1}$School of Information Technology, Monash University, Malaysia Campus\\
$^{2}$School of Electrical and Electronic Engineering, Nanyang Technological University \\
$^{3}$Statistics Program, King Abdullah University of Science and Technology
}

\begin{document}

% Official spconf.sty uses 10 pt by default. ICASSP permits fonts no smaller
% than 9 pt. Uncomment the next line only if you intentionally want 9 pt text.
% \ninept

\maketitle
%\fontsize{9.5}{11.6}\selectfont
% -----------------------------------------------------------------------------
% ABSTRACT: approximately 100--150 words; use the same abstract text in the
% online submission form. Keep it concise and self-contained.
% -----------------------------------------------------------------------------
\begin{abstract}
Electroencephalography (EEG)-based visual decoding and reconstruction aims to recover visual information from non-invasive brain signals. But there are challenges due to the complex temporal-frequency characteristics of EEG signals and the difficulty of discrete denoising framework in mapping neural activity to meaningful visual representations. Here, we propose E2I-Flow, an EEG-to-image decoding and reconstruction framework that combines a novel EEG encoder for time-frequency representation learning from EEG features, and conditional flow matching for visual decoding. A stationary-wavelet frequency branch and a temporal-mixing branch are integrated through bidirectional cross-attention and adaptive fusion, with the resulting EEG representations aligned to ground-truth image embeddings using contrastive and mean-squared-error loss supervisions. The aligned EEG representation is then used as a condition for a Transformer-based flow-matching model to predict the corresponding visual embedding for image synthesis. Experimental results show strong zero-shot EEG-to-image retrieval and improved visual reconstruction performance, outperforming several state-of-the-art baselines.

% validating the effectiveness of the proposed framework.
\end{abstract}


% -----------------------------------------------------------------------------
% 1. INTRODUCTION
% ICASSP requires a discussion of how the contribution relates to prior work.
% This can be integrated here or placed in a separate section.
% -----------------------------------------------------------------------------
\section{Introduction}
\label{sec:introduction}

Decoding visual information from human brain activity enables us to
investigate how visual perception is represented in the brain and has important 
potential for non-invasive brain-computer interfaces (BCIs) 
\cite{Kamitani2005,Gao2021}. Among the commonly used neural recording modalities, EEG is particularly attractive due to its 
non-invasive acquisition, high temporal resolution, relatively low cost, and 
portability \cite{Li2024ATM}. Recent studies have advanced from visual object recognition to zero-shot recognition and image retrieval settings
%enabling finer-grained decoding of perceived visual content 
\cite{Song2024NICE,Li2024ATM}. And a more challenging objective is to reconstruct the visual content itself from EEG signals, which recently has become a promising direction in EEG-based visual decoding \cite{Singh2023EEG2Image,Dai2026AVDE}.

Compared with recognition and retrieval, EEG-to-image reconstruction requires decoded neural information to visually meaningful content. 
%Early studies explored adversarial generation, including ThoughtViz and EEG2IMAGE, which employed GAN-based frameworks to synthesize images from EEG signals \cite{Tirupattur2018ThoughtViz,Singh2023EEG2Image}. 
Early studies explored GAN-based frameworks for synthesizing images from EEG signals \cite{Singh2023EEG2Image}. The emergence of diffusion models provide new solutions to this task. For instance, DM-RE2I incorporated EEG-derived semantic information into a denoising diffusion process \cite{Zeng2023DMRE2I}, whereas DreamDiffusion, BrainVis, and ATM explored alternative ways of
conditioning diffusion models to reconstruct images from learned EEG representations \cite{Bai2024DreamDiffusion,Fu2025BrainVis,Li2024ATM}. In addition to diffusion models, AVDE introduced an autoregressive next-scale prediction approach for EEG-based visual reconstruction \cite{Dai2026AVDE}. Nonetheless, a flow-based generative framework remains largely unexplored for EEG-to-image reconstruction.

%The effectiveness of EEG visual decoding largely depends on how well the encoder captures the complex temporal-frequency characteristics of EEG signals. 
EEG visual decoding depends strongly on how well the encoder captures temporal-frequency characteristics. Existing EEG encoders relied commonly on convolutional or Transformer-based architectures \cite{Schirrmeister2017,Song2023EEGConformer}, whereas more recent studies incorporated frequency-domain information to complement temporal EEG representations \cite{Zhang2025CategoryAware,choi2025visual,Fu2025BrainVis}. Recent studies have also explored multi-scale wavelet modeling and flow-based generation \cite{Tew2026DSFM}. However, jointly modeling these representations within a unified EEG encoder remains limited, particularly for preserving multi-scale information while capturing temporal dynamics.

\begin{figure*}[t]
\centering
\includegraphics[width=0.98\textwidth]{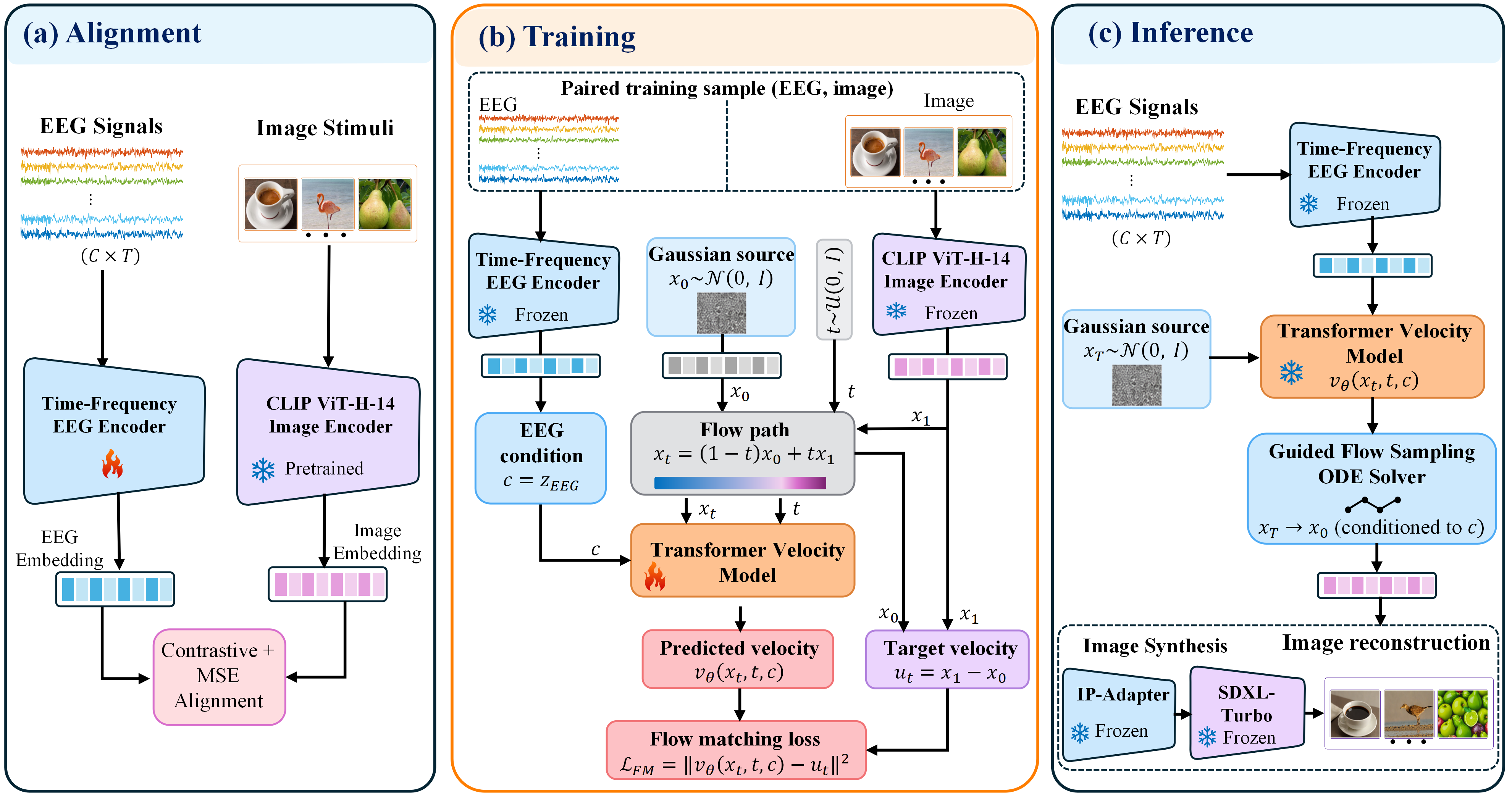}
% \caption{(a) The alignment framework for EEG signals and image stimulus. EEG signals go through the time-frequency EEG encoder to obtain the EEG embeddings, and the image stimulus are input to a pretrained image encoder to generate the image embeddings. (b) Architecture of training part for image reconstruction guided by the EEG condition. It's mainly depend on the flow matching path and transformer velocity model. (c) Inference step for image reconstruction based on the pretrained EEG encoder, transformer velocity model and image synthesis pipeline. }
\vspace{-3mm}
\caption{Overview of E2I-Flow. (a) EEG--image representation alignment using the time-frequency EEG encoder and a pretrained image encoder. (b) Conditional flow matching training, where the aligned EEG representation conditions the Transformer velocity model. (c) Inference through guided flow sampling followed by IP-Adapter and SDXL-Turbo for image synthesis.}
\label{fig:framework}
\vspace{-4mm}
\end{figure*}
To address these limitations, we propose E2I-Flow, an EEG-based visual decoding and reconstruction framework that combines a unified time-frequency EEG representation with conditional flow matching (CFM). The framework first captures temporal and spectral EEG features and guide the flow matching process in the visual embedding space for image reconstruction. %The framework first captures complementary temporal and frequency information from EEG signals and then uses the resulting embeddings to condition a flow matching process that transports samples toward the visual embedding space for image reconstruction. 
Our contributions are threefold: (1) a novel time-frequency EEG encoder that integrates stationary wavelet transform for multi-scale spectral modeling; (2) a bidirectional cross-attention and adaptive fusion mechanism that jointly captures spectral and temporal representations; and (3) a conditional flow matching framework that formulates EEG-to-image reconstruction as a continuous-time latent dynamical process, 
% to visual embeddings for image reconstruction.
providing an alternative to diffusion and autoregressive approaches.
\section{Methods}
\label{sec:method}
% Fig.~\ref{fig:framework} shows an overall framework of E2I-Flow . 
% First, a time-frequency EEG encoder extract EEG embeddings and align with pretrained visual embeddings in a shared representation space, as detailed in Section 2.1 and 2.2, respectively. 
% Next, Section 2.3 presents a CFM model learns that translate the learned EEG embeddings into the corresponding latent visual embeddings. 
% Finally, the generated visual embeddings are fed into an IP-Adapter as a conditional inputs to guide the pretrained SDXL pipeline to decode images. 
% The following subsection describe the proposed time-frequency EEG encoder, the CFM module, and the overall training objective.

\subsection{Time-Frequency EEG Encoder}
\label{ssec:time-frequency}
\begin{figure}[t]
\centering
\includegraphics[width=\columnwidth]{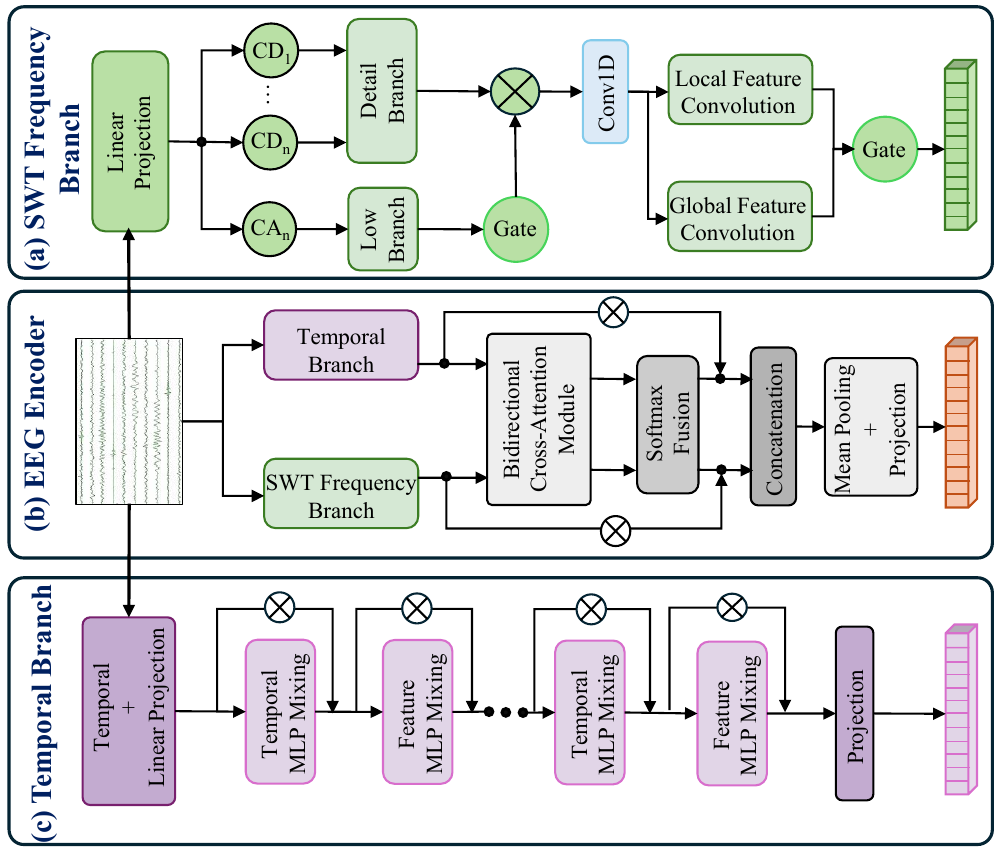}
% \caption{(b)The middle block is the main architecture of the time-frequency EEG encoder. \textcolor{red}{(a) The swt branch extracts frequency information from processed EEG signals. (c)} The temporal branch employs temporal MLP mixing and feature MLP mixing layers to extract informative temporal representations.}
\vspace{-8mm}
\caption{Architecture of time-frequency EEG encoder: (a) SWT frequency branch, (b) Bidirectional cross-attention with adaptive fusion, and (c) Temporal MLP-mixing branch.}
\label{fig:encoder}
\vspace{-4mm}
\end{figure}

Fig.~\ref{fig:framework} shows an overall framework of E2I-Flow. 
% \textcolor{red}{HH: need to emphasize how SWT better than DWT and fourier here.}
% To learn high-quality representations of EEG data, it is crucial to consider the frequency and temporal properties of EEG signals, therefore, we first 
First, to capture complementary spectral and temporal characteristics of EEG signals, we propose a time-frequency EEG encoder, consisting of three main components% stationary wavelet transform (SWT) frequency branch, a temporal branch, and a bidirectional cross-attention and adaptive fusion module
, as illustrated in Fig.~\ref{fig:encoder}. %The EEG representation is decomposed by the stationary wavelet transform (SWT) into approximation and multi-scale detail components. SWT avoids downsampling, preserves temporal alignment across scales, and provides multiresolution time-frequency localization. 
The stationary wavelet transform (SWT) decomposes the EEG signal into the approximation and multi-scale detail components without downsampling \cite{Nason1995SWT}, thereby preserving temporal alignment across scales while providing multi-resolution time-frequency representations.
The approximation and detail frequency components are processed by Conv1D-based branches, with adaptive softmax weighting used to aggregate the detail representations. A convolutional module captures localized patterns, whereas a transformer-based global network models long-range dependencies. Learnable gates adaptively balance the approximation-detail and local-global representations, allowing the fusion weights to vary according to the input EEG sample, producing $F_{\mathrm{SWT}}\in\mathbb{R}^{B\times63\times256}$.
%The temporal branch complements the frequency representation by modeling EEG temporal dynamics using stacked mixing layers. Temporal mixing aggregates information across time points, whereas feature mixing performs nonlinear transformation across latent feature dimensions. Residual connections are employed in both stages to preserve the original representation.
The temporal branch models EEG temporal dynamics using stacked temporal mixing and feature mixing layers with residual connections, producing temporal features that are projected into 63 summary tokens, corresponding to the number of EEG channels, yielding $T_{\mathrm{TS}}\in\mathbb{R}^{B\times63\times256}$. The frequency and temporal representations then interact through bidirectional cross-attention, producing 
$H_{ft}=\mathrm{CrossAttn}(F_{\mathrm{SWT}},T_{\mathrm{TS}})$ and $H_{tf}=\mathrm{CrossAttn}(T_{\mathrm{TS}},F_{\mathrm{SWT}})$. In the frequency-to-temporal direction, $F_{\mathrm{SWT}}$ serves as query, while $T_{\mathrm{TS}}$ as keys and values; the reverse direction exchanges these roles. The two outputs are adaptively combined through softmax fusion to obtain $H_{\mathrm{fusion}}$. Residual connections are subsequently applied to the frequency and temporal branch outputs, yielding $U_f=F_{\mathrm{SWT}}+H_{\mathrm{fusion}}$ and $U_t=T_{\mathrm{TS}}+H_{\mathrm{fusion}}$, which are then concatenated, mean-pooled, and projected to obtain the final EEG embedding \(z_{\mathrm{EEG}}\in\mathbb{R}^{B\times1024}\).
%\vspace{-4pt}
\subsection{EEG-Image Representation Alignment}
\label{ssec:overview}
%\noindent\textbf{EEG-Image Representation Alignment}\\[-2pt]
% We maximize agreement between corresponding EEG-image pairs while minimizing similarity between non-corresponding pairs. Let $B$ denotes the batch size, $C$ the number of EEG channels, $T$ the number of sampled time points, and $H$ and $W$ the height and width of the image, respectively. Our aim is to learn EEG representations $e\in \mathbb{R}^{B \times d}$ from the processed EEG data $E \in \mathbb{R}^{B \times C \times T}$, aligned with the corresponding image embeddings $z \in \mathbb{R}^{B \times d}$ obtained from ground truth images $\qquad I \in \mathbb{R}^{B \times H \times W}$, based on the pretrained CLIP ViT-H-14 image encoder. The cross-modal alignment is optimized using the standard cross-entropy loss:
% We maximize agreement between corresponding EEG--image pairs while reducing similarity between non-corresponding pairs. Let $B$ denote the batch size, $C$ the number of EEG channels, $T$ the number of sampled time points, and $H$ and $W$ the image height and width, respectively. Given processed EEG signals $E\in\mathbb{R}^{B\times C\times T}$, the encoder produces EEG embeddings $e\in\mathbb{R}^{B\times d}$. Corresponding ground-truth images $I\in\mathbb{R}^{B\times H\times W}$ are encoded using the pretrained CLIP ViT-H-14 image encoder to obtain visual embeddings $z\in\mathbb{R}^{B\times d}$. Cross-modal alignment is optimized using a symmetric contrastive objective:
Denote the EEG signals to be $E\in\mathbb{R}^{B\times C\times T}$, where $B$ denote the batch size, $C$ the number of EEG channels. The encoder produces embeddings $e\in\mathbb{R}^{B\times d}$. Corresponding images are encoded by the pretrained CLIP image encoder \cite{Radford2021CLIP} to obtain visual embeddings $z\in\mathbb{R}^{B\times d}$. We align paired EEG and image representations using a symmetric contrastive objective:
\begin{equation}
\begin{aligned}
\mathcal{L}_{\mathrm{CLIP}}
= -\frac{1}{B}\sum_{i=1}^{B}
\Bigg(
&\log
\frac{
\exp\left(s(e_i,z_i)/\tau\right)
}{
\sum_{j=1}^{B}\exp\left(s(e_i,z_j)/\tau\right)
}
\\
&+
\log
\frac{
\exp\left(s(e_i,z_i)/\tau\right)
}{
\sum_{k=1}^{B}\exp\left(s(e_k,z_i)/\tau\right)
}
\Bigg).
\end{aligned}
\end{equation}
where $s$ denotes cosine similarity and $\tau$ is a learned temperature parameter that controls the sharpness of the distribution. This contrastive objective aligns the EEG and image representations in a shared embedding space.

To complement the contrastive objective with point-wise supervision, we add the mean squared error (MSE), which measures the distance between the EEG embeddings and their corresponding image embeddings:
\begin{equation}
\mathcal{L}_{align}
=
\lambda \mathcal{L}_{CLIP}
+
(1-\lambda)\mathcal{L}_{MSE}
\end{equation}
% where $\lambda$ balances the contributions of the two objectives. By combining contrastive alignment with direct regression, resulting a more reliable EEG--image correspondence.
where $\lambda$ balances the two objectives. Combining contrastive alignment with direct regression encourages both discriminative cross-modal correspondence and point-wise agreement between paired EEG and image representations.

% The bundled file figures/example_figure.pdf comes from the official ICASSP
% template package. Replace it with your own PDF/PNG/JPG figure and change the
% filename below. PDF is preferable for vector architecture diagrams.

% For a two-column-wide architecture figure, use figure* instead:
%

\subsection{Conditional Flow Matching Image Reconstruction}
\label{ssec:reconstruction}
We reconstruct visual stimuli from the aligned EEG representation using conditional flow matching \cite{Lipman2023FlowMatching}, which learns a continuous velocity field that transports samples from a simple source distribution toward the target visual-embedding distribution. Specifically, let
$c=z_{\mathrm{EEG}}\in\mathbb{R}^{B\times1024}$
and
$x_1=z_{\mathrm{image}}\in\mathbb{R}^{B\times1024}$
denote the EEG condition and ground-truth CLIP image embedding, respectively. During training, a Gaussian source
$x_0\sim\mathcal{N}(0,I)$
and a timestep
$t\sim U(0,1)$
are sampled to construct the intermediate state
$x_t=(1-t)x_0+t x_1$
and the corresponding target velocity
$u_t=x_1-x_0$
along the flow path. %The EEG condition $c$, intermediate state $x_t$, and temporal information $t$ are then introduced into the transformer velocity model, where positional and token-type embeddings are added to form joint representations, yielding the predicted velocity $v_\theta(x_t,t,c)$.
The EEG condition $c$, intermediate state $x_t$, and timestep $t$ are provided to the Transformer velocity model, together with positional and token-type embeddings, to predict the conditional velocity $v_\theta(x_t,t,c)$.

The model is optimized using the conditional Flow Matching objective
\begin{equation}
%\begin{aligned}
\mathcal{L}_{\mathrm{CFM}}
=
\left\|
v_\theta(x_t,t,c)-u_t
\right\|_2^2 \\
%&=
%\left\|
%v_\theta(x_t,t,c)-(x_1-x_0)
%\right\|_2^2 .
%\end{aligned}
\end{equation}
At inference time, the trained Transformer evaluates the conditional and unconditional velocities $v_\theta(x_t,t,c)$ and $v_\theta(x_t,t,\varnothing)$, respectively, which are combined using classifier-free guidance (CFG) \cite{Ho2022ClassifierFreeGuidance}.
\begin{equation}
\hat{v}(x_t,t,c)
=
v_\theta(x_t,t,\varnothing)
+
s\left[
v_\theta(x_t,t,c)
-
v_\theta(x_t,t,\varnothing)
\right]
\end{equation}
where $\hat{v}$ denotes the guided velocity, 
%conditional and unconditional velocity predictions are $v_\theta(x_t,t,c)$ and $v_\theta(x_t,t,\varnothing)$, respectively. 
$\varnothing$ is a zero vector used in place of the EEG condition embedding, and $s$ is the guidance scale. The guided velocity $\hat{v}(x_t,t,c)$ is used by the ODE solver to iteratively update the current state from $t=0$ to $t=1$, progressively transforming the initial Gaussian state into the predicted image embedding
$\hat{z}_{\mathrm{image}}\in\mathbb{R}^{B\times1024}$.  Finally, the predicted visual representation
$\hat{z}_{\mathrm{image}}$ is converted into the reconstructed RGB image using pretrained IP-Adapter \cite{Ye2023IPAdapter} and SDXL-Turbo \cite{Sauer2023ADD}.
\section{Experimental Results}
\label{sec:experiments}
% ============================================================
% Table 1: Full-width within-subject retrieval
% ============================================================

%\begin{figure}[t]
%    \centering
%    \includegraphics[width=\columnwidth]%{figures/wavelet_level_top1_ablation.png}
%    \vspace{-4pt}
%    \caption{Top-1 retrieval performance under different wavelet decomposition levels.}
%    \label{fig:wavelet_level_ablation}
%\end{figure}

%\begin{figure}[t]
%    \centering
%    \includegraphics[
%        width=\columnwidth,
%        trim=0 6 0 0,
%        clip
%    ]{figures/wavelet_level_top1_ablation.png}
%    \caption{Top-1 retrieval performance under different wavelet decomposition levels.}
%    \label{fig:wavelet_level_ablation}
%\end{figure}

% \subsection{Experimental Setup}
% \label{ssec:setup}
\noindent\textbf{Dataset.}
We conduct experiments on the THINGS-EEG benchmark
\cite{Gifford2022THINGS}, following the standard split used in prior EEG visual decoding studies \cite{Song2024NICE,Li2024ATM,Dai2026AVDE}. The training set contains 1,654 concepts with 10 images per concept, while the test set contains 200 unseen concepts with one image per concept. Each training image was presented four times per subject and each test image 80 times. EEG signals were recorded from 63 channels at 1 KHz. We extracted 0--1,000 millisecond epochs, downsampled to 200 Hz, and, for evaluation, repeated trials corresponding to the same image were averaged before EEG encoding.

\begin{table*}[t]
\centering
%\vspace{-4mm}
\caption{Within-subject 200-way zero-shot EEG-to-image retrieval performance
(\%) on the THINGS-EEG dataset. T1 and T5 denote Top-1 and Top-5 accuracy,
respectively. Results are averaged over five random seeds.}
\label{tab:retrieval}

% ===== SIZE CONTROL START =====
\footnotesize
\setlength{\tabcolsep}{3.00pt}
\renewcommand{\arraystretch}{1.50}
% ===== SIZE CONTROL END =====

\begin{tabular}{@{}l*{22}{c}@{}}
\toprule

\multirow{2}{*}{\textbf{Method}}
& \multicolumn{2}{c}{\textbf{Sub-01}}
& \multicolumn{2}{c}{\textbf{Sub-02}}
& \multicolumn{2}{c}{\textbf{Sub-03}}
& \multicolumn{2}{c}{\textbf{Sub-04}}
& \multicolumn{2}{c}{\textbf{Sub-05}}
& \multicolumn{2}{c}{\textbf{Sub-06}}
& \multicolumn{2}{c}{\textbf{Sub-07}}
& \multicolumn{2}{c}{\textbf{Sub-08}}
& \multicolumn{2}{c}{\textbf{Sub-09}}
& \multicolumn{2}{c}{\textbf{Sub-10}}
& \multicolumn{2}{c}{\textbf{Mean}}
\\

& T1 & T5
& T1 & T5
& T1 & T5
& T1 & T5
& T1 & T5
& T1 & T5
& T1 & T5
& T1 & T5
& T1 & T5
& T1 & T5
& T1 & T5
\\

\midrule

Conformer (2023)~\cite{Song2023EEGConformer}
& 9.5  & 26.1
& 10.8 & 27.4
& 14.2 & 30.6
& 15.5 & 37.1
& 8.8  & 19.8
& 8.1  & 28.0
& 12.2 & 31.8
& 16.4 & 40.5
& 10.9 & 29.8
& 15.1 & 39.2
& 12.2 & 31.0
\\

NICE (2024)~\cite{Song2024NICE}
& 20.1 & 47.9
& 19.2 & 36.9
& 21.2 & 53.8
& 22.4 & 50.4
& 14.4 & 31.6
& 26.1 & 56.3
& 26.9 & 55.7
& 38.2 & 67.4
& 23.4 & 53.2
& 29.8 & 58.6
& 24.2 & 51.2
\\

ATM (2024)~\cite{Li2024ATM}
& 23.2 & 51.2
& 18.8 & 43.2
& 27.3 & 57.0
& 28.0 & 54.1
& 16.8 & 39.5
& 28.0 & 59.2
& 26.8 & 53.7
& 39.3 & 71.5
& 24.5 & 51.2
& 37.2 & 67.7
& 26.9 & 54.8
\\

VDF (2025)~\cite{choi2025visual}
& 20.4 & 49.0
& 16.5 & 49.0
& 17.5 & 51.0
& 25.5 & \textbf{58.0}
& 13.0 & 43.0
& 23.0 & 54.5
& 19.5 & 48.0
& 24.0 & 55.0
& 18.0 & 50.0
& 20.5 & 49.5
& 19.8 & 50.7
\\

AVDE (2026)~\cite{Dai2026AVDE}
& \textbf{25.0} & \textbf{55.2}
& 24.1 & 51.0
& 27.5 & 58.6
& \textbf{29.8} & 54.7
& \textbf{25.4} & 50.3
& \textbf{33.5} & 60.3
& 27.4 & 55.2
& 41.7 & 71.3
& 26.1 & 52.1
& \textbf{39.5} & \textbf{73.0}
& 30.0 & 58.2
\\

\midrule

\textbf{E2I-Flow (Ours)}
& 23.4 & 48.6
& \textbf{27.7} & \textbf{57.6}
& \textbf{32.5} & \textbf{61.3}
& 28.5 & 57.6
& 22.9 & \textbf{51.0}
& 30.1 & \textbf{62.1}
& \textbf{29.1} & \textbf{61.0}
& \textbf{47.0} & \textbf{75.7}
& \textbf{31.7} & \textbf{58.0}
& 37.0 & 67.7
& \textbf{31.0} & \textbf{60.1}
\\

\bottomrule
\end{tabular}
\vspace{-4mm}
\end{table*}
\noindent\textbf{Implementational Details.}
The EEG encoder was trained with AdamW using a batch size of 128 and subject-specific learning rates of $5\times10^{-4}$ and $9\times10^{-5}$, with a maximum of 50 epochs and early stopping patience of 10. CLIP ViT-H-14 is used to extract 1024-dimensional image embeddings. %The alignment loss uses equal weights for the contrastive and MSE objectives ($\lambda=0.5$), and the SWT branch adopts db1 with two decomposition levels. 
We set $\lambda=0.5$ in the alignment objective and use db1 with two SWT decomposition levels. 
The conditional flow-matching model was trained for 150 epochs using Adam with a learning rate of $10^{-3}$ and a batch size of 1024. At inference, the learned flow was integrated over $t\in[0, 1]$ using a 50-step midpoint ODE solver with a classifier-free guidance scale of 14.
\begin{table}[t]
\centering
\setlength{\belowcaptionskip}{4pt}
\caption{Average quantitative comparison of EEG-based visual reconstruction over all 10 subjects.}
\vspace{-1mm}
\label{tab:reconstruction_average}

\footnotesize
\setlength{\tabcolsep}{2.0pt}
\renewcommand{\arraystretch}{1.08}

\begin{tabular}{@{}lccccccc@{}}
\toprule
& \multicolumn{2}{c}{Low-level}
& \multicolumn{5}{c}{High-level} \\
\cmidrule(lr){2-3}
\cmidrule(lr){4-8}

Method
& PixCorr $\uparrow$
& SSIM $\uparrow$
& Alex(2) $\uparrow$
& Alex(5) $\uparrow$
& Inc. $\uparrow$
& CLIP $\uparrow$
& SwAV $\downarrow$ \\
\midrule

ATM~\cite{Li2024ATM}
& 0.130
& 0.339
& 0.740
& 0.822
& 0.685
& 0.716
& 0.599 \\

AVDE~\cite{Dai2026AVDE}
& 0.147
& \textbf{0.366}
& 0.766
& 0.835
& \textbf{0.724}
& 0.747
& 0.587 \\

\textbf{Ours}
& \textbf{0.155}
& \textbf{0.366}
& \textbf{0.771}
& \textbf{0.849}
& 0.721
& \textbf{0.787}
& \textbf{0.565} \\

\bottomrule
\end{tabular}
\vspace{-4mm}
\end{table}
%\vspace{10pt}
\begin{figure}[t]
\centering
\includegraphics[width=\columnwidth]{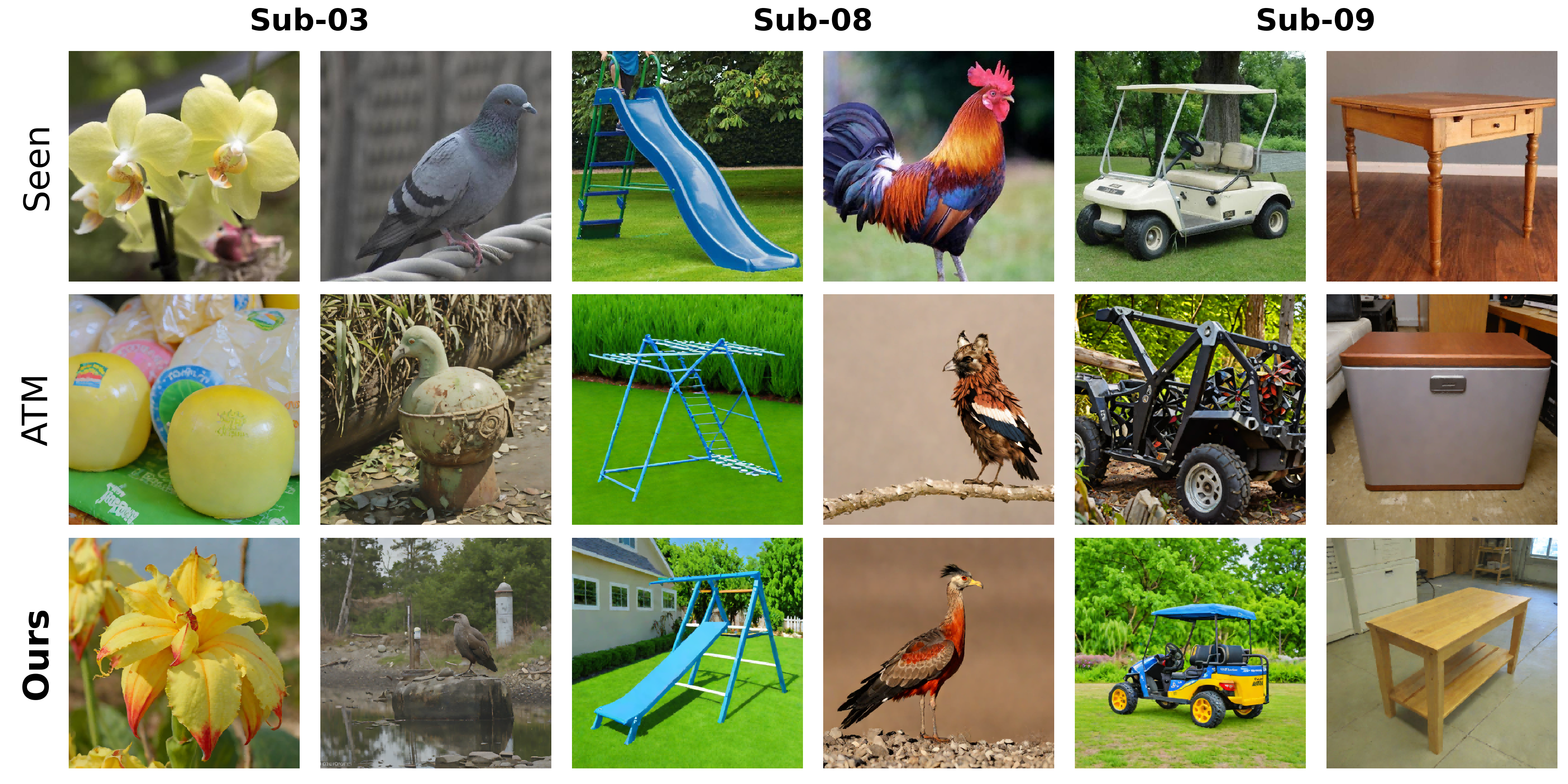}
% \caption{Qualitative comparison of visual reconstruction across multiple subjects. The reconstructions produced by our method preserve clearer object content and visual details.}
\vspace{-7mm}
\caption{Qualitative comparison of EEG-based image reconstruction across multiple subjects using ATM, E2I-Flow, and the corresponding ground-truth stimuli.}
\label{fig:compa_fig}
\vspace{-4mm}
\end{figure}
\begin{table}[t]
\centering
% \caption{Ablation study of the proposed EEG encoder on selected subjects. T1 and T5 denote Top-1 and Top-5 retrieval accuracy (\%), respectively. All variants use the same training protocol with a fixed random seed.}
\setlength{\belowcaptionskip}{4pt}
\caption{E2I-Flow ablation on selected subjects. T1 and T5 denote Top-1 and Top-5 retrieval accuracy (\%), respectively.}
\vspace{-1mm}
\label{tab:encoder_ablation}
\footnotesize
\setlength{\tabcolsep}{2.0pt}
\renewcommand{\arraystretch}{1.10}
\begin{tabular}{@{}l*{8}{c}@{}}
\toprule
& \multicolumn{2}{c}{Sub-01}
& \multicolumn{2}{c}{Sub-02}
& \multicolumn{2}{c}{Sub-09}
& \multicolumn{2}{c}{Sub-10} \\
\cmidrule(lr){2-3}
\cmidrule(lr){4-5}
\cmidrule(lr){6-7}
\cmidrule(lr){8-9}
Framework
& T1 & T5
& T1 & T5
& T1 & T5
& T1 & T5 \\
\midrule

SWT Frequency Only
& 17.0 & 40.0
& 24.5 & 52.5
& 23.5 & 52.5
& 33.5 & 63.0 \\

Temporal Only
& 18.0 & 39.5
& 25.5 & 53.5
& 26.0 & 49.5
& \textbf{35.0} & 65.0 \\

SWT + Temporal + MLP Fusion
& 20.0 & \textbf{41.5}
& 24.5 & 54.5
& \textbf{29.0} & 52.5
& 33.5 & 61.5 \\

\textbf{E2I-Flow (Ours)}
& \textbf{21.0} & \textbf{41.5}
& \textbf{27.0} & \textbf{58.5}
& 28.0 & \textbf{56.5}
& \textbf{35.0} & \textbf{65.5} \\

\bottomrule
\end{tabular}
\vspace{-3mm}
\end{table}
\begin{table}[t]
\centering
%\vspace{6pt}
\setlength{\belowcaptionskip}{4pt}
\caption{Hyperparameter analysis of wavelet decomposition settings. Top-1 retrieval accuracy (\%) is reported for different wavelet families and decomposition levels.}
\vspace{-1mm}
\footnotesize
\setlength{\tabcolsep}{5pt}
\label{tab:wavelet_hyperparameter}
\begin{tabular}{ccccccc}
\toprule
Wavelet & Level 2 & Level 3 & Level 4 & Level 5 & Level 6 & Level 7 \\
\midrule
db1 & 48.0 & 43.5 & 47.5 & 43.0 & 44.5 & 44.0 \\
db4 & 46.5 & 44.5 & 47.5 & 43.0 & 43.5 & 42.5 \\
db8 & 43.0 & 43.5 & 46.5 & 39.5 & 43.0 & 42.0 \\
\bottomrule
\end{tabular}
\vspace{-4mm}
\end{table}
% \subsection{Main results}

\noindent\textbf{Main results.} For EEG-to-image retrieval, we evaluate the matching scores between the predicted EEG representations and the CLIP image embeddings of the 200 unseen test concepts. Retrieval performance is reported using Top-1 and Top-5 accuracy. For image reconstruction, we follow prior work \cite{Scotti2023MindEye,Li2024ATM}. %Low-level similarity is measured using PixCorr and SSIM, while high-level similarity is evaluated using AlexNet-2 and AlexNet-5 \cite{Krizhevsky2012AlexNet}, Inception \cite{Szegedy2015Inception}, CLIP, and SwAV \cite{Caron2020SwAV}.
Low-level reconstruction quality is evaluated using PixCorr and SSIM, while high-level perceptual quality is assessed using AlexNet-2, AlexNet-5~\cite{Krizhevsky2012AlexNet}, Inception~\cite{Szegedy2015Inception}, CLIP, and SwAV~\cite{Caron2020SwAV} based metrics.

As shown in Table~\ref{tab:retrieval}, %E2I-Flow achieves competitive within-subject 200-way zero-shot retrieval performance across the evaluated subjects, with strong Top-1 and Top-5 results on most subjects. 
E2I-Flow achieves the highest mean Top-1 and Top-5 retrieval accuracies of 31.0\% and 60.1\%, respectively, compared with 30.0\% and 58.2\% for the best competing method, AVDE.
For visual reconstruction, Table~\ref{tab:reconstruction_average} shows that E2I-Flow achieves the best average performance on five of the seven metrics, including PixCorr, AlexNet-2, AlexNet-5, CLIP, and SwAV, while remaining highly competitive on SSIM and Inception. In particular, the improvement in CLIP similarity suggests stronger preservation of high-level visual semantics. Consistent with the quantitative results, the qualitative comparison in Fig.~\ref{fig:compa_fig} shows that our method
recovers clearer object content and finer visual details.

% \subsection{Ablation and analysis}
% As shown in Table~\ref{tab:encoder_ablation}, using either the SWT frequency branch or the temporal branch alone generally degrades retrieval performance, while replacing bidirectional cross-attention and adaptive fusion with MLP fusion produces less consistent results. These results support the complementary roles of frequency and temporal modeling and the effectiveness of their adaptive interaction. 
\noindent\textbf{Ablation Study.} Table~\ref{tab:encoder_ablation} evaluates the contributions of the frequency, temporal, and fusion components. In general, the complete encoder achieves the best performance. Using either the SWT or temporal branch alone generally reduces performance, while replacing bidirectional cross-attention and adaptive fusion with MLP fusion yields less consistent improvements. This attests to the benefits of leveraging on the complementary information of frequency and temporal modeling, and the adaptive cross-modal interaction between the two representations. Table~\ref{tab:wavelet_hyperparameter} further shows that increasing SWT decomposition depth does not consistently improve retrieval performance; the final configuration of E2I-Flow therefore uses db1 with two decomposition levels.%therefore, the db1 wavelet family with two decomposition levels is adopted in the final model.
% -----------------------------------------------------------------------------
% 4. CONCLUSION
% -----------------------------------------------------------------------------
\section{Conclusion}
\label{sec:conclusion}
% In this paper, we developed E2I-Flow model,  achieving competitive retrieval performance and improved visual reconstruction quality. The experimental results demonstrate its effectiveness for EEG-based visual decoding and reconstruction.
We developed the E2I-Flow, a novel EEG-to-image reconstruction framework that combines wavelet-temporal representation learning with conditional flow matching. The proposed encoder integrates multi-scale SWT features and temporal representations through bidirectional cross-attention and adaptive fusion, while the conditional flow model maps aligned EEG representations toward the visual embedding space for reconstruction. Experiments on THINGS-EEG show improved mean zero-shot retrieval performance and stronger reconstruction quality across most evaluated perceptual metrics, supporting the effectiveness of the proposed framework. In future work, we plan to extend E2I-Flow from static image reconstruction to EEG-to-video generation by characterizing dynamic visual information.

% -----------------------------------------------------------------------------
% REFERENCES
% The official IEEEbib.bst file is included unchanged. The two bibliography
% entries supplied here are the same demonstration entries distributed in the
% official ICASSP template package; replace them with real references.
%
% The optional fifth page may contain ONLY references, funding acknowledgements,
% and a Compliance with Ethical Standards statement when required by ICASSP.
% Do not place technical text, figures, or tables on page 5.
% -----------------------------------------------------------------------------
% Retain the two official demonstration references so the preview shows the
% bibliography style. Remove this line once you add citations in the manuscript.
%\nocite{Lamp86,C2}
%\bibliographystyle{IEEEbib}
%\bibliography{strings,refs}
{\fontsize{9.7}{11.2}\selectfont
\bibliographystyle{IEEEbib}
\bibliography{strings,refs}

@article{Kamitani2005,
  author  = {Yukiyasu Kamitani and Frank Tong},
  title   = {Decoding the Visual and Subjective Contents of the Human Brain},
  journal = {Nature Neuroscience},
  volume  = {8},
  number  = {5},
  pages   = {679--685},
  year    = {2005},
  doi     = {10.1038/nn1444}
}

@article{Gao2021,
  author  = {Xiaorong Gao and Yijun Wang and Xiaogang Chen and Shangkai Gao},
  title   = {Interface, Interaction, and Intelligence in Generalized Brain--Computer Interfaces},
  journal = {Trends in Cognitive Sciences},
  volume  = {25},
  number  = {8},
  pages   = {671--684},
  year    = {2021},
  doi     = {10.1016/j.tics.2021.04.003}
}

@inproceedings{Song2024NICE,
  author    = {Yonghao Song and Bingchuan Liu and others},
  title     = {Decoding Natural Images from {EEG} for Object Recognition},
  booktitle = {International Conference on Learning Representations},
  year      = {2024}
}

@inproceedings{Li2024ATM,
  author    = {Dongyang Li and Chen Wei and Shiying Li and Jiachen Zou and Quanying Liu},
  title     = {Visual Decoding and Reconstruction via {EEG} Embeddings with Guided Diffusion},
  booktitle = {Advances in Neural Information Processing Systems},
  volume    = {37},
  year      = {2024},
  doi       = {10.52202/079017-3266}
}

@inproceedings{Singh2023EEG2Image,
  author    = {Prajwal Singh and Pankaj Pandey and others},
  title     = {{EEG2IMAGE}: Image Reconstruction from {EEG} Brain Signals},
  booktitle = {2023 IEEE International Conference on Acoustics, Speech and Signal Processing (ICASSP)},
  pages     = {1--5},
  year      = {2023},
  doi       = {10.1109/ICASSP49357.2023.10096587}
}

@inproceedings{Dai2026AVDE,
  author    = {Sicheng Dai and Hongwang Xiao and Shan Yu and Qiwei Ye},
  title     = {Autoregressive Visual Decoding from {EEG} Signals},
  booktitle = {International Conference on Learning Representations},
  year      = {2026}
}

@article{Zeng2023DMRE2I,
  author  = {Hong Zeng and Nianzhang Xia and others},
  title   = {{DM-RE2I}: A Framework Based on Diffusion Model for the Reconstruction from {EEG} to Image},
  journal = {Biomedical Signal Processing and Control},
  volume  = {86},
  pages   = {105125},
  year    = {2023},
  doi     = {10.1016/j.bspc.2023.105125}
}

@inproceedings{Fu2025BrainVis,
  author    = {Honghao Fu and Hao Wang and Jing Jih Chin and Zhiqi Shen},
  title     = {{BrainVis}: Exploring the Bridge between Brain and Visual Signals via Image Reconstruction},
  booktitle = {2025 IEEE International Conference on Acoustics, Speech and Signal Processing (ICASSP)},
  pages     = {1--5},
  year      = {2025},
  doi       = {10.1109/ICASSP49660.2025.10889805}
}

@article{Schirrmeister2017,
  author  = {Robin Tibor Schirrmeister and Jost Tobias Springenberg and others},
  title   = {Deep Learning with Convolutional Neural Networks for {EEG} Decoding and Visualization},
  journal = {Human Brain Mapping},
  volume  = {38},
  number  = {11},
  pages   = {5391--5420},
  year    = {2017},
  doi     = {10.1002/hbm.23730}
}

@article{Song2023EEGConformer,
  author  = {Yonghao Song and Qingqing Zheng and Bingchuan Liu and Xiaorong Gao},
  title   = {{EEG} Conformer: Convolutional Transformer for {EEG} Decoding and Visualization},
  journal = {IEEE Transactions on Neural Systems and Rehabilitation Engineering},
  volume  = {31},
  pages   = {710--719},
  year    = {2023},
  doi     = {10.1109/TNSRE.2022.3230250}
}

@inproceedings{Zhang2025CategoryAware,
  author    = {Enshang Zhang and Zhicheng Zhang and Takashi Hanakawa},
  title     = {Category-Aware {EEG} Image Generation Based on Wavelet Transform and Contrast Semantic Loss},
  booktitle = {Proceedings of the Thirty-Fourth International Joint Conference on Artificial Intelligence (IJCAI)},
  pages     = {7922--7930},
  year      = {2025},
  doi       = {10.24963/ijcai.2025/881}
}

@article{Gifford2022THINGS,
  author  = {Alessandro T. Gifford and Kshitij Dwivedi and Gemma Roig and Radoslaw M. Cichy},
  title   = {A Large and Rich EEG Dataset for Modeling Human Visual Object Recognition},
  journal = {NeuroImage},
  volume  = {264},
  pages   = {119754},
  year    = {2022},
  doi     = {10.1016/j.neuroimage.2022.119754}
}

@inproceedings{Scotti2023MindEye,
  author    = {Paul S. Scotti and Atmadeep Banerjee and others},
  title     = {Reconstructing the Mind's Eye: {fMRI}-to-Image with Contrastive Learning and Diffusion Priors},
  booktitle = {Advances in Neural Information Processing Systems},
  volume    = {36},
  pages     = {24705--24728},
  year      = {2023}
}

@inproceedings{Krizhevsky2012AlexNet,
  author    = {Alex Krizhevsky and Ilya Sutskever and Geoffrey E. Hinton},
  title     = {{ImageNet} Classification with Deep Convolutional Neural Networks},
  booktitle = {Advances in Neural Information Processing Systems},
  volume    = {25},
  year      = {2012}
}

@inproceedings{Szegedy2015Inception,
  author    = {Christian Szegedy and Wei Liu and others},
  title     = {Going Deeper with Convolutions},
  booktitle = {Proceedings of the IEEE Conference on Computer Vision and Pattern Recognition},
  pages     = {1--9},
  year      = {2015},
  doi       = {10.1109/CVPR.2015.7298594}
}

@inproceedings{Caron2020SwAV,
  author    = {Mathilde Caron and Ishan Misra and others},
  title     = {Unsupervised Learning of Visual Features by Contrasting Cluster Assignments},
  booktitle = {Advances in Neural Information Processing Systems},
  volume    = {33},
  year      = {2020}
}

@inproceedings{Bai2024DreamDiffusion,
  author    = {Yunpeng Bai and Xintao Wang and others},
  title     = {DreamDiffusion: High-Quality EEG-to-Image Generation with Temporal Masked Signal Modeling and CLIP Alignment},
  booktitle = {Proceedings of the European Conference on Computer Vision},
  year      = {2024},
  doi       = {10.1007/978-3-031-72751-1_27}
}

@inproceedings{Tew2026DSFM,
  author    = {Hwa Hui Tew and Junn Yong Loo and others},
  title     = {Functional MRI Time Series Generation via Wavelet-Based Image Transform and Spectral Flow Matching for Brain Disorder Identification},
  booktitle = {International Conference on Learning Representations},
  year      = {2026}
}

@inproceedings{choi2025visual,
  title     = {Visual Decoding Using a Learnable Wavelet-Based Spatial-Spectral-Temporal EEG Embedding},
  author    = {Choi, Yebin and Kim, Jun-Mo and others},
  booktitle = {2025 13th International Conference on Brain-Computer Interface (BCI)},
  pages     = {1--5},
  year      = {2025},
  organization = {IEEE}
}

@inproceedings{Radford2021CLIP,
  author    = {Alec Radford and Jong Wook Kim and others},
  title     = {Learning Transferable Visual Models From Natural Language Supervision},
  booktitle = {Proceedings of the 38th International Conference on Machine Learning},
  volume    = {139},
  pages     = {8748--8763},
  year      = {2021}
}

@inproceedings{Lipman2023FlowMatching,
  author    = {Yaron Lipman and Ricky T. Q. Chen and others},
  title     = {Flow Matching for Generative Modeling},
  booktitle = {International Conference on Learning Representations},
  year      = {2023}
}

@article{Ho2022ClassifierFreeGuidance,
  author  = {Jonathan Ho and Tim Salimans},
  title   = {Classifier-Free Diffusion Guidance},
  journal = {arXiv preprint arXiv:2207.12598},
  year    = {2022}
}

@article{Ye2023IPAdapter,
  author  = {Hu Ye and Jun Zhang and others},
  title   = {{IP-Adapter}: Text Compatible Image Prompt Adapter for Text-to-Image Diffusion Models},
  journal = {arXiv preprint arXiv:2308.06721},
  year    = {2023}
}

@article{Sauer2023ADD,
  author  = {Axel Sauer and Dominik Lorenz and Andreas Blattmann and Robin Rombach},
  title   = {Adversarial Diffusion Distillation},
  journal = {arXiv preprint arXiv:2311.17042},
  year    = {2023}
}

@incollection{Nason1995SWT,
  author    = {Guy P. Nason and Bernard W. Silverman},
  title     = {The Stationary Wavelet Transform and Some Statistical Applications},
  booktitle = {Wavelets and Statistics},
  series    = {Lecture Notes in Statistics},
  volume    = {103},
  pages     = {281--299},
  publisher = {Springer},
  year      = {1995},
  doi       = {10.1007/978-1-4612-2544-7_17}
}
}
\end{document}